\documentclass[letterpaper, 10 pt, conference]{ieeeconf}  % Comment this line out if you need a4paper

\IEEEoverridecommandlockouts                              % This command is only needed if 
\usepackage{graphicx} 
\usepackage{xcolor}
\usepackage{cite}
\usepackage{amsmath,amssymb,amsfonts}
\usepackage{algorithmic}
\usepackage{textcomp}
\usepackage{comment}

\usepackage{tikz}
\usetikzlibrary{plotmarks}
\newcommand\marksymbol[2]{\tikz[#2,scale=1.2]\pgfuseplotmark{#1};}

\newcommand{\um}[0]{{µm }}

\newcommand{\nm}[0]{{N·m }}

\title{\LARGE \bf
Johnsen-Rahbek Capstan Clutch: A High Torque Electrostatic Clutch
}

\author{Timothy E. Amish$^{1}$, Jeffrey T. Auletta$^{2}$, Chad C. Kessens$^{2}$, Joshua R. Smith$^{1,3}$, and Jeffrey I. Lipton$^{4*}$% <-this % stops a space
\thanks{$^{1}$ Dept of Electrical and Computer Engineering, University of Washington, Seattle, WA, 98195 USA }%
\thanks{$^{2}$ US Army Research Directorate, DEVCOM Army Research Laboratory, Aberdeen Proving Ground, MD 21005 USA }%
\thanks{$^{3}$ Paul G Allen School of Computer Science and Engineering, University of Washington, Seattle, WA, 98195 USA }%
\thanks{$^{4}$ Mechanical and Industrial Engineering Department of Northeastern University, Boston, MA, 02115 USA }%
\thanks{$^{*}$ {\tt\small j.lipton@northeastern.edu}}%
\vspace{-.35cm}
}

\begin{document}

\maketitle
\thispagestyle{empty}
\pagestyle{empty}

%%%%%%%%%%%%%%%%%%%%%%%%%%%%%%%%%%%%%%%%%%%%%%%%%%%%%%%%%%%%%%%%%%%%%%%%%%%%%%%%
\begin{abstract}
%following https://www.nature.com/documents/nature-summary-paragraph.pdf
%Basic Intro
In many robotic systems, the holding state consumes power, limits operating time, and increases operating costs. 
Electrostatic clutches have the potential to improve robotic performance by generating holding torques with low power consumption.
% Specific program
A key limitation of electrostatic clutches has been their low specific shear stresses which restrict generated holding torque, limiting many applications.
% Here we so
Here we show how combining the Johnsen-Rahbek (JR) effect with the exponential tension scaling capstan effect can produce clutches with the highest specific shear stress in the literature. 
Our system generated 31.3~N/cm\textsuperscript{2} sheer stress and a total holding torque of 7.1~N·m while consuming only 2.5~mW/cm\textsuperscript{2} at 500~V. We demonstrate a theoretical model of an electrostatic adhesive capstan clutch and demonstrate how large angle ($\theta>2\pi$) designs increase efficiency over planar or small angle ($\theta<\pi$) clutch designs. We also report the first unfilled polymeric material, polybenzimidazole (PBI), to exhibit the JR-effect.
\end{abstract}

%%%%%%%%%%%%%%%%%%%%%%%%%%%%%%%%%%%%%%%%%%%%%%%%%%%%%%%%%%%%%%%%%%%%%%%%%%%%%%%%

\section{INTRODUCTION}

% I edited the 1st 3 paragraphs here
Clutches are critical in many robotic systems, particularly in applications where Size Weight and Power (SWaP) are key constraints. %[S.Collins - Efficient bipedal], aerial robots [Ollero - control techniques for aerial,Stirling - Energy efficient indoor search by warms] and exoskeletons [Pratt- The roboknee, Dollar -Lower extremity exoskeletons]. 
%Generally, mobile high degrees-of-freedom (DoF) robots need to be low SWaP to have reasonable operating times. 
To produce robots with many degrees of freedom (DoF), low SWaP components with significant holding forces and torques are crucial. This paper describes a new capstan-based electrostatic clutch design that produces higher torque scaling capabilities than conventional electrostatic brake-based designs, while maintaining the attractive low SWaP properties of earlier electrostatic brakes.  The capstan effect is synergistically utilized with Johnsen-Rahbek (JR) type electrostatic adhesion \cite{Johnsen-1923} leading to the name, JR-effect driven capstan clutch (JRCC).

Conventional fully actuated robots employ one actuator per joint \cite{Tsai-1995-tendon,He-2019-underactuated}. Using one motor per joint tends to be heavy, particularly in high DoF designs, such as hands \cite{Billard-2019-trends}. Clutches are typically integrated into robotic systems to reduce weight and power in one of two ways: mechanical multiplexing or a braking system. In mechanical multiplexing, power is routed between multiple outputs, which saves weight and power by reducing the number of actuators. As a braking mechanism, clutches block unwanted motion consuming power \cite{Plooij-2015-lock}. Capstan-based clutches are especially useful devices that take a small input tension and exponentially scale an output tension by wrapping a flexible line around a shaft \cite{Stuart-1961-Capstan,Cacucciolo-2022-Peeling}.

% \begin{figure}[t]
%  \centering
%  \includegraphics[width=0.9\columnwidth]{Figures/Arm_Experiment.png}
%  \vspace{-3.5mm}
%  \caption{a) JR-effect driven capstan clutch (JRCC) holding at torque of 2.16 N·m (2.0 kg weight, 110 mm moment arm) with an input voltage of 1000 V. b) Clutch disengaged (0 V applied)}.
%  \label{fig:Hero}
%  \vspace{-0.8cm}
% \end{figure}

\begin{figure}[t]
 \centering
 \includegraphics[width=0.9\columnwidth]{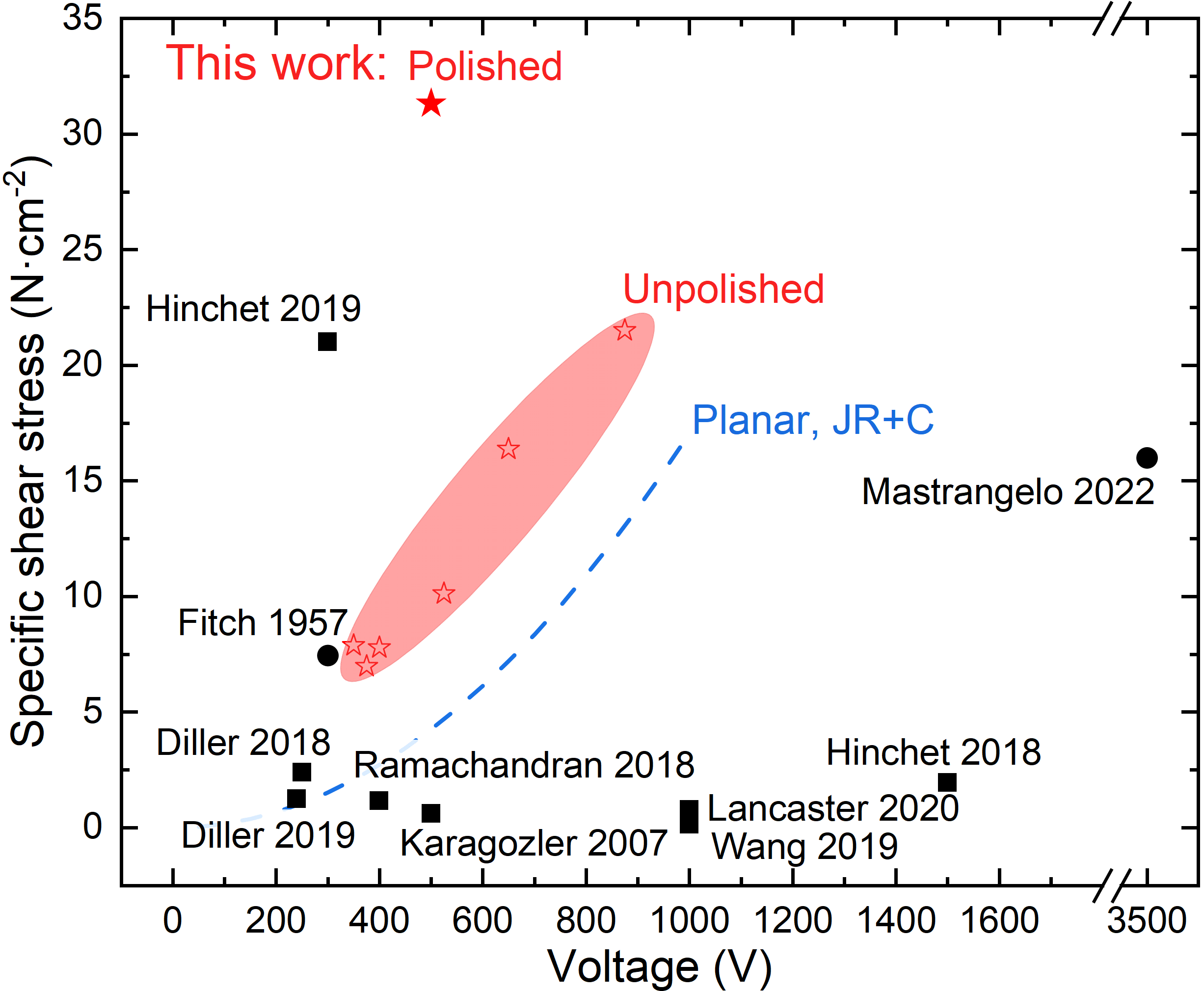}
 \vspace{-3mm}
 \caption[.]{
 {Comparison of JRCC against other reported electrostatic clutches (\marksymbol{square*}{black}~=~planar and \marksymbol{*}{black}~=~curved). The highlighted region is the maximum observed for various wrap angles of a JRCC with a 25.4 µm band. The star point is for the JRCC with a 76.2 µm band demonstrating the highest recorded specific shear stress. The dashed line is a comparison to an equivalent planar clutch.}}
 \label{fig:comparison}
 \vspace{-0.75cm}
\end{figure}

Electrostatic clutches are based on the attraction force between two electrodes at different voltages separated by a dielectric. They can be implemented in a light, thin, and low power fashion for SWaP constrained robots. Compared to traditional clutches, electrostatic clutches are more power efficient, lighter-weight \cite{Guo-2020-tech-review,Diller-2018-DesignParam} and have even been shown to have a specific tension capability better than biological muscle \cite{Patrick-2020-underactuated}. 

Despite the advantages of both capstan and electrostatic clutches, the combination of their governing effects have had only limited demonstration, with application angles under $\pi$ radians \cite{Wei-2023-SoftRobotic-Multiplexing,Fitch-1957-IBM-Development,Johnsen-1923}. Due to the exponential nature of the capstan effect, such devices improve holding torque and power efficiency as the number of wraps increases. The primary metrics for an electrostatic clutch's efficiency are the specific stress (N/cm$^{2}$), and specific power (mW/cm$^{2}$). Specific shear stress is commonly reported in the literature as it represents the efficiency of the material at generating forces.
 
In this work, we contribute a design for a JR-effect driven multi-wrap capstan clutch (JRCC) utilizing electrostatic adhesion. We demonstrate how incorporating a multi-wrap capstan design increases the electrostatic clutch's specific stress and specific power. %Fig.~\ref{fig:Hero} is a demonstration of a JRCC with a 2.16 N·m holding torque. In the 0 V state, the JRCC allows the center shaft to move freely. %resists no motion of the center shaft.
We instantiate the design using two different bands, a thin stainless steel band and a thicker band that has been polished. We use these bands to investigate the effects that yield stress and surface roughness have on our design. As seen in Fig.~\ref{fig:comparison}, we generate the highest specific shear stress of any clutch in the literature. We present a model for such systems and demonstrate agreement with the model using experimental data. 
%Due to our use of the capstan effect, the power consumption of our clutch is much less than that required for an equivalent linear clutch design. With the proposed capstan electrostatic clutch, the voltage and active area work to scale an exponential term greatly increasing the holding torque possible. 
 
In this paper we:

\begin{itemize}
    \item Design and fabricate a JR-effect driven, multi-wrap capstan clutch that can generate 7.1~N·m of holding torque and the highest specific shear stress in the literature,
    \item Demonstrate the first unfilled polymer, polybenzimidazole (PBI), to exhibit the JR effect,
    \item Develop and validate a model for electrostatic capstan clutches and analyze the design space,
    \item Experimentally quantify the advantage of JRCC designs over planar designs, and
    \item Explore the effect of band thickness, surface finish, and wrap angle on clutch efficiency and holding torque.
    
\end{itemize}

\section{Background}
\subsection{Electrostatic adhesion}

Electrostatic adhesive (EA) devices are found in a wide variety of applications, from semiconductor chucking systems \cite{Sogard-2007-Chucks} to actuation systems \cite{Rothemund-2021-Hasel} and robotic end effectors \cite{Lancaster-2022-effector,Guo-2020-tech-review}.  In a typical EA device, an electrode is adhered to one side of a dielectric material. A second electrode acts as a braking surface between itself and the open surface of the dielectric. Depending on the volume resistivity of the dielectric, there are two different regimes of electroadhesion: Coulombic and Johnsen-Rahbek (JR) \cite{Kanno-2006-JR-Prediction,Levine-2021-electroprogrammable-stiffness}. Both Coulombic and JR forces and equations are described in Fig.~\ref{fig:theory}.

%Fixed some errors in the description of the cicuit models
Most EA devices utilize a dielectric with high volume resistivity, $\rho$~$\approx$~$10^{13-18}$~$\Omega\cdot$cm, that corresponds to Coulombic forces only. The Coulombic EA force may be modeled as a series combination of two parallel RC networks. One network corresponds to the air gap between the dielectric and the electrode. The other corresponds to the capacitor formed by the dielectric itself. The resulting normal force is predicted to scale with the square of the applied voltage as shown in Fig.~\ref{fig:theory}, where $A$ is the apparent contact area, $\varepsilon_0$ the permittivity of free space, $d$ the dielectric thickness, $g$ the gap distance, and $\varepsilon_d$ and $\varepsilon_g$ are the dielectric and air gap permittivity \cite{Sogard-2007-Chucks,Persson-2018-Coulomb-Theory-Derivation-Dependency,Chen-2017-comparison-low-voltage,Strong-1970-electro-display,Nakamura-2017-modeling-control}.

%Since everyone is here...thinking of stating something like...Both Coulomb and JR-force may be modeled using the same circuit (whatever descruption sounds best/we agree upon)? Then give differnce based on volume resistivity and subsequent descriptions. 

%Under an applied voltage, the Coulombic EA force may be modeled as two parallel RC circuits in series due to the formation of an air gap between the dielectric and electrode.

%In contrast to a Coulombic EA, 

%For a dielectric with $\rho$~$\approx$~$10^{9-13}$~$\Omega\cdot$cm, an additional attractive force is present, termed the Johnsen-Rahbek force (JR) \cite{Kanno-2006-JR-Prediction}, and may also be modeled as two parallel RC circuits in series \cite{Nakamura-2017-modeling-control}. Here, due to the dielectric's lower volume resistivity and migration of charge towards the electrode surface, much of the applied voltage is present at the micron-sized gap between the interfaces. Due to constricting surface asperities, the effective gap voltage is approximated as the applied voltage since the contact resistance is much greater than the bulk resistance of the dielectric. This produces a strong EA force, shown in Fig.~\ref{fig:theory}, which only depends on the gap distance and dielectric constant (in this case air) \cite{Johnsen-1923,Fitch-1957-IBM-Development,Kanno-2006-JR-Prediction,Kanno-2003-generation-mechanism,Sogard-2007-Chucks,Nakamura-2017-modeling-control,Watanabe-1993-resistivity,Balakrishnan-1950-oldJR}. Note: the total EA normal force in our JRCC is given by the sum of the Coulomb and JR force. 

%% Josh Version
For a dielectric with $\rho$~$\approx$~$10^{9-13}$~$\Omega\cdot$cm (low for an insulator), an additional attractive force is present, termed the Johnsen-Rahbek force (JR) \cite{Kanno-2006-JR-Prediction}, which may also be modeled as a series combination of two parallel RC circuits \cite{Nakamura-2017-modeling-control}. Due to the dielectric's relatively low volume resistivity and migration of charge towards the electrode surface, most of the applied voltage appears at the micron-sized gap between the interfaces. In a Coulombic EA, the force is limited by the dielectric thickness; in a JR EA, there is effectively an induced capacitor plate on the open surface of the dielectric. The gap between the induced plate and its corresponding electrode is therefore much smaller than the dielectric thickness, limited mainly by surface roughness. This results in a much larger EA force than conventional Coulombic designs. As shown in Fig.~\ref{fig:theory}, the JR force only depends on the voltage, gap distance and dielectric constant (in this case air) \cite{Johnsen-1923,Fitch-1957-IBM-Development,Kanno-2006-JR-Prediction,Kanno-2003-generation-mechanism,Sogard-2007-Chucks,Nakamura-2017-modeling-control,Watanabe-1993-resistivity,Balakrishnan-1950-oldJR}. Finally, the total EA normal force in our JRCC is given by the sum of the Coulomb and JR forces. 

%Here, the system may also be considered as a two capacitors in series. where the voltage drop occurs across the interfacial gap . due to charge migration within the dielectric toward the electrode surface, 

%As a result, approximately the applied voltage is present at the micron-sized gap between the interfaces, https://www.overleaf.com/project/64decfa1fba8452d1f877b16

\subsection{Capstan Effect}
A capstan is a passive, self-amplifying brake where a cable is wrapped around a shaft (also called a capstan) \cite{Samset-1985-Winch}. As the cable is tightened around the shaft by tensioning or affixing one end, the frictional force holds the cable in place \cite{Plooij-2015-lock}. The capstan's advantage comes from the holding tension scaling exponentially with the total angle swept around the shaft \cite{Stuart-1961-Capstan}. Generally, capstan winches are used in marine or industrial applications where a human operator can hold entire ships or large equipment in place with little input tension. Since the basic operation of a capstan relies on tension, they naturally lend themselves to tendon driven applications, functioning as a brake or clutch for control \cite{Kang-2012-nonElectric-capstan-clutch,Sinclair-2019-Capstan-VR}.

\subsection{Electrostatic Clutches}

Electrostatic Clutches (ESCs) date back to the early work of Johnsen and Rahbek in 1923, and possibly earlier to an 1875 patent issued to Elisha Gray \cite{Johnsen-1923,Fitch-1957-IBM-Development}. 
%The output of Johnsen and Rahbek's clutch was attached to a diaphragm producing a sound wave modulated by electrically controlling the friction between a band and central shaft\cite{Johnsen-1923}. 
Even though the capstan effect is likely present in Johnsen and Rahbek's design, it is not stated or referenced. The capstan effect was also found to significantly improve the performance of EA soft grippers on curved surfaces, although wrap angles~$>$$\pi/2$ were not explored\cite{Cacucciolo-2022-curved-model,Cacucciolo-2019-curved-with-metric,Cacucciolo-2023-curved-zipping}. A feasibility analysis on an EA clutch design with a curved braking was conducted but critically did not use the capstan effect\cite{Detailleur-2021-Feasibility-Analysis}.

Here, we explicitly exploit the capstan and JR effects and derive a model for the output tension \eqref{gov-eqn} of an ESC. For applications in SWaP constrained robots, ESCs are an attractive solution as they generally consume only a few milliwatts of power \cite{Guo-2020-tech-review}, can be engaged and disengaged on the order of milliseconds making them valuable for fast control of robots \cite{Hinchet-2020-High-force}, and, although not explored in this work, have self-sensing capabilities \cite{Guo-2018-sensing}.  

ESCs are typically constructed in a planar design. Flexible ESCs can conform to different objects and retain shape with applied voltage \cite{Guo-2018-sensing,Wang-2019-Variable-Stiffness}.  A high force, flexible ESC was integrated into a glove used to lock hand position for VR \cite{Hinchet-2020-High-force}. A planar design allows ESCs to be stacked, such as in an ankle support application \cite{Diller-2016-clutch}. In our previous work, a stack of ESCs were used in a finger-inspired robot gripper \cite{Lancaster-2022-effector}, and in a mechanically multiplexed ten DoF tentacle robot using one ESC per DoF. By toggling the clutches on and off, one motor was able to actuate the entire tentacle robot \cite{Patrick-2020-underactuated}. A JRCC does lack this stacking characteristic but leverages a more powerful exponentially scaling holding tension at higher wrap angles. Moreover, revolute joints are most common in robotics, and a planar design needs additional mechanisms to be integrated, whereas a JRCC integrates directly. 

%Here, our work tackles the basic design and modeling of a novel capstan enabled electrostatic clutch that has torque advantages over the traditional planar design.

% Probably don't need Oragami paper. By placing clutches into a grid and selectively activating specific clutches, different structures are formed \cite{Karagozler-2007-Oragami}.

%%%%%%%%%%%%%%%%%%%%%%%%%%%%%%%%%%%%%%%%%%%%%%%%%%%%%%%%%%%%%%%%%%%%%%
\section{Modeling Capstan and JR Effects}
\begin{figure}[t]
 \centering
 \includegraphics[width=0.9\columnwidth]{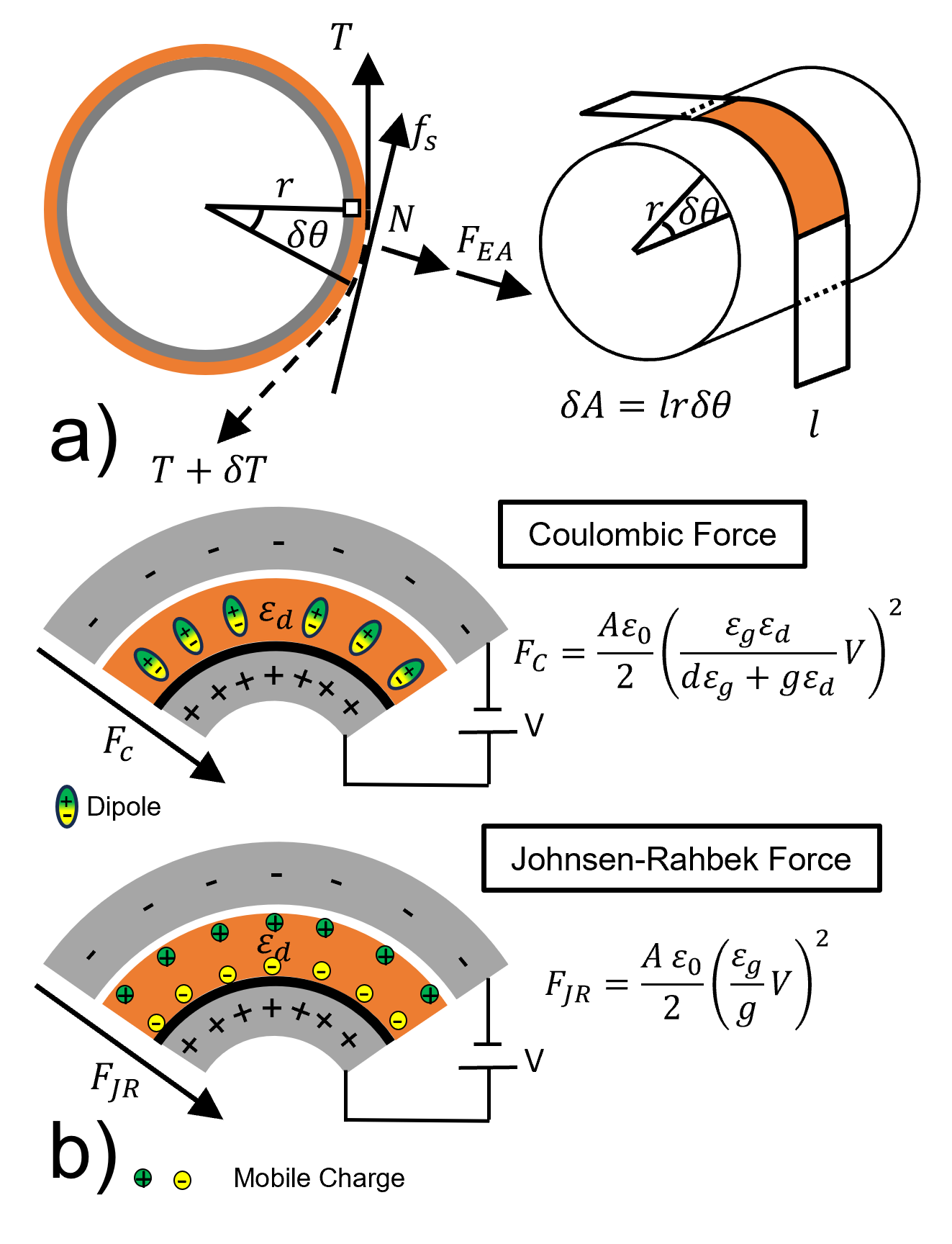}
 \vspace{-3.5mm}
 \caption{a) The electrostatic capstan effect adds an electrically driven force component to the Normal forces in a standard capstan drive. Unlike the holding forces of a standard capstan, the electroactive force is a function of area. b) Using a PBI material, both JR and Coulombic electroadhesion is present.}
 \label{fig:theory}
 \vspace{-0.5cm}
\end{figure}

The device described in this work takes advantage of the exponential nature of the capstan effect and combines it with JR and Coulombic electrostatic attraction as shown in Fig. \ref{fig:theory}. The traditional capstan effect model as a function of the holding tension and wrap angle is 
\begin{equation}
T_{load} = T_{hold}e^{\mu\theta_{total}}\label{capstan}
\end{equation}
where $\mu$ is the coefficient of friction between the capstan band and the dielectric, and $\theta$ is the total angle swept by the band around the center shaft.

The derivation of the electrostatic capstan effect also relies on the same assumptions that are used to derive \eqref{capstan}. The rope (stainless steel band in our case) must be on the verge of slipping, meaning $T_{load}$ is at a maximum. The assumed ideal rope must be compliant and non-elastic. %This last notes lends to why their are two versions of the JRCC. A characteristic design to validate theory with a thin electrode that acts similar to the ideal rope. The optimized JRCC uses a thicker electrode that is much more rigid but can withstand larger applied torques. 

In a JRCC the wrapped electrode is attracted to the center capstan, producing an additional term to the normal force $F_{ea}$, Fig.~\ref{fig:theory}. This electrostatic force is a function of a variety of material parameters and the contact area. The area is the only parameter that is a function of small angle $\delta A = lr\delta \theta$, where $l$ is the width of the band and $r$ is the radius of the center shaft. We therefore isolate the dependence on $\theta$ and collapse the other geometric and material terms into a single constant, $\alpha$
\begin{equation}
F_{EA} = \alpha\cdot\delta\theta\label{diff-EA}
\end{equation}
Using \eqref{diff-EA} for $F_{EA}$ as shown in Fig. \ref{fig:theory} and integrating over appropriate bounds produces \eqref{capstan-general-es}, which is the holding tension for a generic electrostatic capstan clutch.

\begin{equation}
T_{load} = T_{hold}e^{\mu\theta}+\alpha(e^{\mu\theta}-1)\label{capstan-general-es}
\end{equation}

The PBI dielectric used in this work acts as a JR active material, meaning there are two categories of electrostatic adhesion at work: 1) Coulomb force and 2) JR force, shown in Fig.~\ref{fig:theory}. These sources of electrostatic adhesion can be combined and substituted for alpha in equation \eqref{capstan-general-es} to produce the governing equation \eqref{gov-eqn} evaluated in this paper for the JRCC. 

\vspace{-5.0mm}
\begin{comment}
\begin{equation}
F_{C} = \frac{A}{2}\varepsilon_0 \left(\frac{\varepsilon_g\varepsilon_d}{d\varepsilon_g + g\varepsilon_d}\right)^2 V^2\label{coulomb}
\end{equation}

\begin{equation}
F_{JR} = \frac{A}{2}\varepsilon_0 \left(\frac{\varepsilon_g}{g}\right)^2 V^2\label{JR}
\end{equation}
\end{comment}

\begin{multline}
T_{load}=T_{hold}e^{\mu\theta} +\\ \frac{\varepsilon_0}{2}V^2 lr\left[\left(\frac{\varepsilon_g\varepsilon_d}{d\varepsilon_g + g\varepsilon_d}\right)^2+ \left(\frac{\varepsilon_g}{g}\right)^2\right](e^{\mu\theta}-1)\label{gov-eqn}
\end{multline}

%The capstan effect adds a multiplicative term to the normal force $(e^{\mu\theta}-1)$. 

An equivalent flat area system can be modeled by replacing the term $(e^{\mu\theta}-1)$ with $\theta$ and multiplying by $\mu$ to get 

\begin{equation}
T_{planar} = \mu\frac{\varepsilon_0}{2}V^2 lr\theta\left[\left(\frac{\varepsilon_g\varepsilon_d}{d\varepsilon_g + g\varepsilon_d}\right)^2 + \left(\frac{\varepsilon_g}{g}\right)^2\right]\label{planar-eq}
\end{equation}

Defining a new constant $\beta = \alpha/lr$ that represents the electrostatic constants, we can compute the advantage an electrostatic capstan has by dividing \eqref{gov-eqn} by \eqref{planar-eq}, giving an advantage term of 
\begin{equation}
Adv = \frac{\left(\frac{T_{hold}}{lr\beta}+1\right)e^{\mu\theta}-1}{\mu\theta}\approx\frac{e^{\mu\theta}-1}{\mu\theta} \label{adv-eqn}
\end{equation}

$\frac{T_{hold}}{lr\beta}$ represents the ratio of the holding torque to the electroactive force.
$\frac{T_{hold}}{lr\beta}$ must be $ << 1$ to allow the system to rotate easily when disengaged, and was found to be 0.007 in our JRCC construction. We can then approximate the advantage as shown in \eqref{adv-eqn}. If the advantage number is less than 1, a planar design has an advantage; if it is greater than one, the JRCC design has an advantage. In the limit where theta goes to zero, the advantage number becomes identically 1, and is larger than 1 for all other angles. Therefore, a electrostatic capstan clutch will always perform better than a planar clutch of the same material, area and gap size. The capstan advantage term also informs the mechanical design of the clutch. Increasing $l$ or $r$ does increase the holding forces the clutch can apply, but the exponential advantage of the capstan is a function of $\theta$ alone. %Therefore a design should use as narrow a band and small a shaft diameter possible and increase wrap angle to more efficiently convert electroactive forces into output forces. 

An important design factor for the capstan brake is the yield stress of the band materials. The cross sectional area of the band is $l h$, where $h$ is the thickness of the band. If we assume $T_{hold} << 1$, the yield stress can be found

\begin{equation}
\sigma_{max} =  \frac{r}{h} \beta (e^{\mu\theta}-1)
\label{eqn:stress}
\end{equation}

The stress in the band is not a function of its width, but the output force is a function of width. Therefore, when operating at the maximum limit of the band material, using a wider band can linearly improve performance. 

The torque output is $T_{load}$ multiplied by the radius of the central shaft $r$. Solving for the maximum holding torque as a function of maximum yield stress $\sigma_{max}$ is

\begin{equation}
\tau_{max} = l h r \sigma_{max}
\label{eqn:max-torque}
\end{equation}

Therefore, for a fixed maximum stress of a band material, we can linearly improve the output torque by increasing the band thickness, width, or the radius of the clutch. Accordingly, two versions of the JRCC utilizing different band thicknesses are reported: a thin band for evaluating our model and a thick band for assessing higher holding torque.

The limit of the band material will always be reached by increasing the wrap angle until $T_{load} = T_{max}$. At that point, increasing brake performance depends on selecting a different band material, or linear improvements based on geometric trade-offs.  Increasing the performance of the dielectric material will improve the $\beta$ term and more efficiently convert voltage into holding force. With a higher $\beta$, a smaller radius shaft will generate the same holding torques. 
\vspace{-0.40cm}
\section{Clutch Design}
\begin{figure}[t]
 \centering
 \includegraphics[width=0.9\columnwidth]{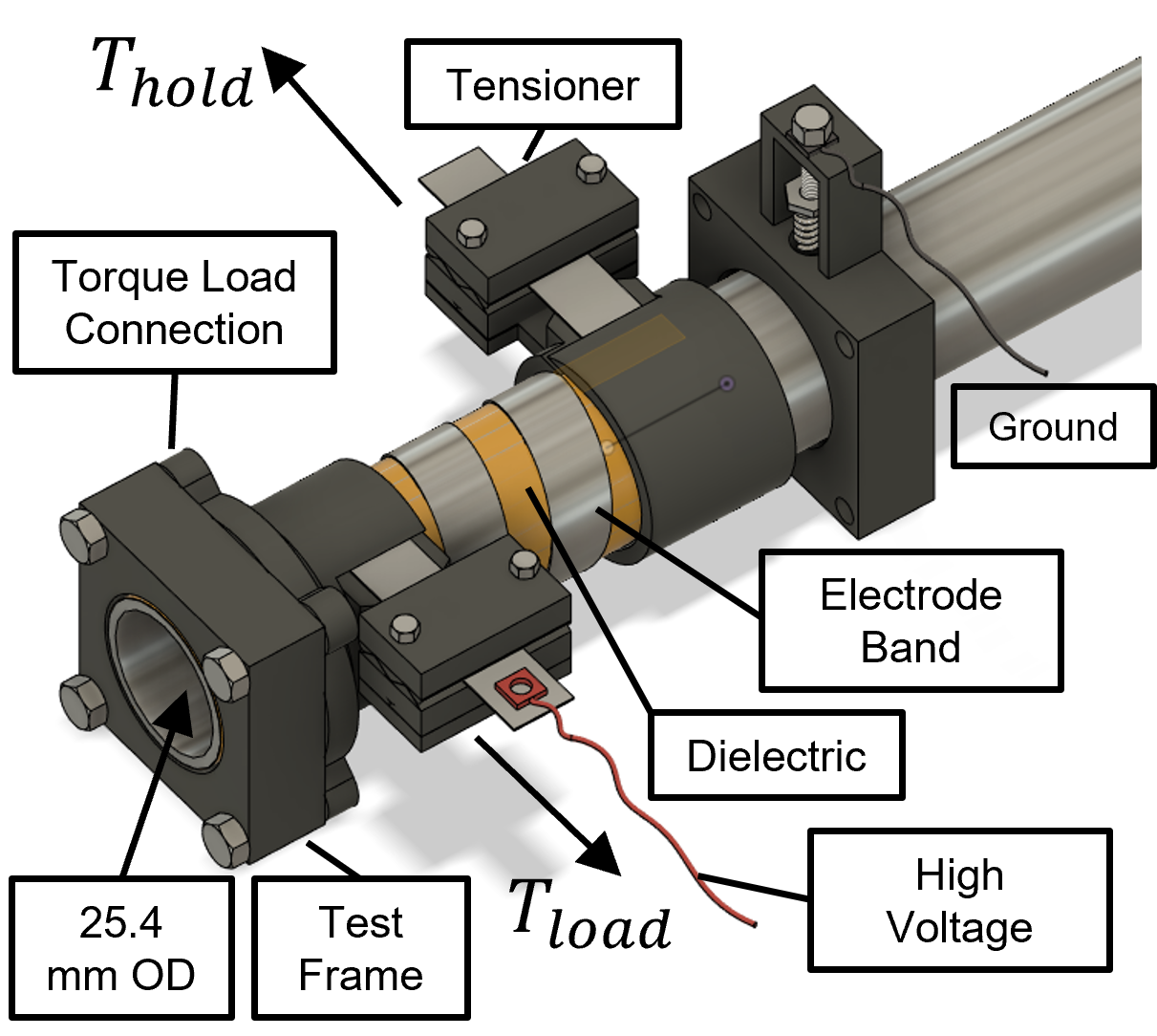}
 \vspace{-3mm}
 \caption{Design for a multi-wrap JR-effect driven capstan clutch (JRCC). The design consists of a stainless steel band wrapped around a PBI dielectric on a 25.4~mm diameter stainless steel shaft. This design can generate up to 7.1~N·m of holding torque.}
 \label{fig:Design}
 \vspace{-0.75cm}
\end{figure}

\vspace{-0.35cm}
\subsection{Clutch Hardware}
 The main body of the electrostatic clutch is 3D printed on a Markforged~X7 using Onyx filament. One end of the clutch acts as the output, and a weight is attached to the other end, acting as the pretension ($T_{hold}$). The band is wrapped around the central stainless steel 25.4~mm diameter shaft at an angle of 10~deg to produce a separation of 3~mm between wraps. A 55~µm film of Polybenzimidazole (PBI) is readily wrapped and adhered to the central shaft using double sided carbon tape. PBI was chosen as it demonstrated both Coulombic and JR electrostatic adhesion. %The polymer was provided by the .

 A clamp holds the band and rides on a bearing to ensure that the band enters and exits the central shaft at a correct angle without twisting. If the shim is twisted, the clutch will not operate effectively since electrostatic adhesion is sensitive to peeling forces \cite{Cacucciolo-2022-Peeling}. To energize the device, a compressed spring connected to ground rides on the central shaft as it rotates, and the band is connected to high voltage. Different wrap angles are easily achieved by adjusting the ends of the clutch. Only the output end of the clutch is fixed while the pretension end rides freely along the shaft. Consequently, the clutch will only resist motion in the same direction as the wrap angle. The active elements weigh 20.17~g. The entire device including the shaft weights 313.2~g. 
% Currently, one of the ends of the clutch is designated the output and the other is the pretension end. The consequence of designating the output and pretension end is the clutch will only resist rotation in the same direction as the wrap angle. 

\subsection{Clutch Bands}
We tested our design with two stainless steel bands made from shim. One band was a 25.4~µm thick, 10~mm wide. This band was not polished. The second band was a thicker 76.2~µm and polished. The band was polished using a felt pad on a Dremel with 1~µm diamond polishing compound. Both thin and thick bands behave similar to a power spring and exert a small but nontrivial torque which must be counterbalanced to ensure contact with the dielectric. Consequently, 5~g and 200~g weights were chosen to pretension the 25.4~µm and 76.2~µm bands, respectively, to produce near 0~N·m holding torque at 0~V.

%Even though a thicker SS band can resist a higher shear stresses, it acts more rigid than a thinner band, meaning it does not abide by the capstan equation as closely. The added rigidity causes the band to act more similar to a torsional spring, this feature can be overcome some by increasing the holding tension. Since some of the holding tension is to bend the band into shape there will be an error in \eqref{gov-eqn} if the pretension value is not lowered to account for this.

% We measured the compoments of the designs and found that 
% 6.8 g for free end
% 9.0 g for stationary end
% 9.83 g for M3x4 bolts
% 3.37 g per clamp/ball bearing assmebly
% 1.0 g for PBI+carbon tape (full width required for 3 wrap)
% 279.5 g for 30.48 cm shaft
% 0.33 g for 1 cm x 15.24 cm shim.

%\begin{table}[]
%\begin{tabular}{lll}
%Item                 & Number & Weight (g) \\ \hline
%Free end             & 1      & 6.8        \\
%Stationary End       & 1      & 9.0        \\
%M3x4 Bolts           & 10     & 0.983      \\
%Clamp Assembly       & 2      & 3.37       \\
%PBI+Carbon Tape      & 1      & 1.0        \\
%Metal Shaft          & 1      & 279.5      \\
%Stainless Steel Shim & 1      & 0.33     
%\end{tabular}
%\label{tab:components}
%\caption{Part count and weights of the clutch}
%\end{table}

\section{Experimental Characterization of JRCC}
\label{sec:ExpCharacterization}
This section describes the experimental setup and validates the proposed model for the JRCC. The power consumption and effect of pretension is discussed. A further comparison to other clutch designs and validation of the advantage factor is also presented.

\subsection{Experimental Setup}

The experimental setup consisted of the JRCC, DYN-200 torque sensor, Trek 10/10B-HS high voltage power supply, and a computer with LabVIEW for data acquisition. The power was calculated from the current and voltage recorded from the Trek 10/10B-HS. An increasing torque was applied to the torque sensor connected to the JRCC until the clutch slipped.  We found that the 25.4~µm thick band would snap at $\approx$~2.5~N·m of torque. In order to evaluate the model of our clutch design, the applied torque was limited to $\approx$~2~N·m. Unless otherwise noted, we used the 25.4~\um band to collect the data. The thick band was used to increase the holding torque maximum by requiring a higher force to snap. We will call a $2\pi$~radian angle one wrap.

\subsection{Validation of the JR-Capstan Model}

\begin{figure}[t]
 \centering
 \includegraphics[width=0.9\columnwidth]{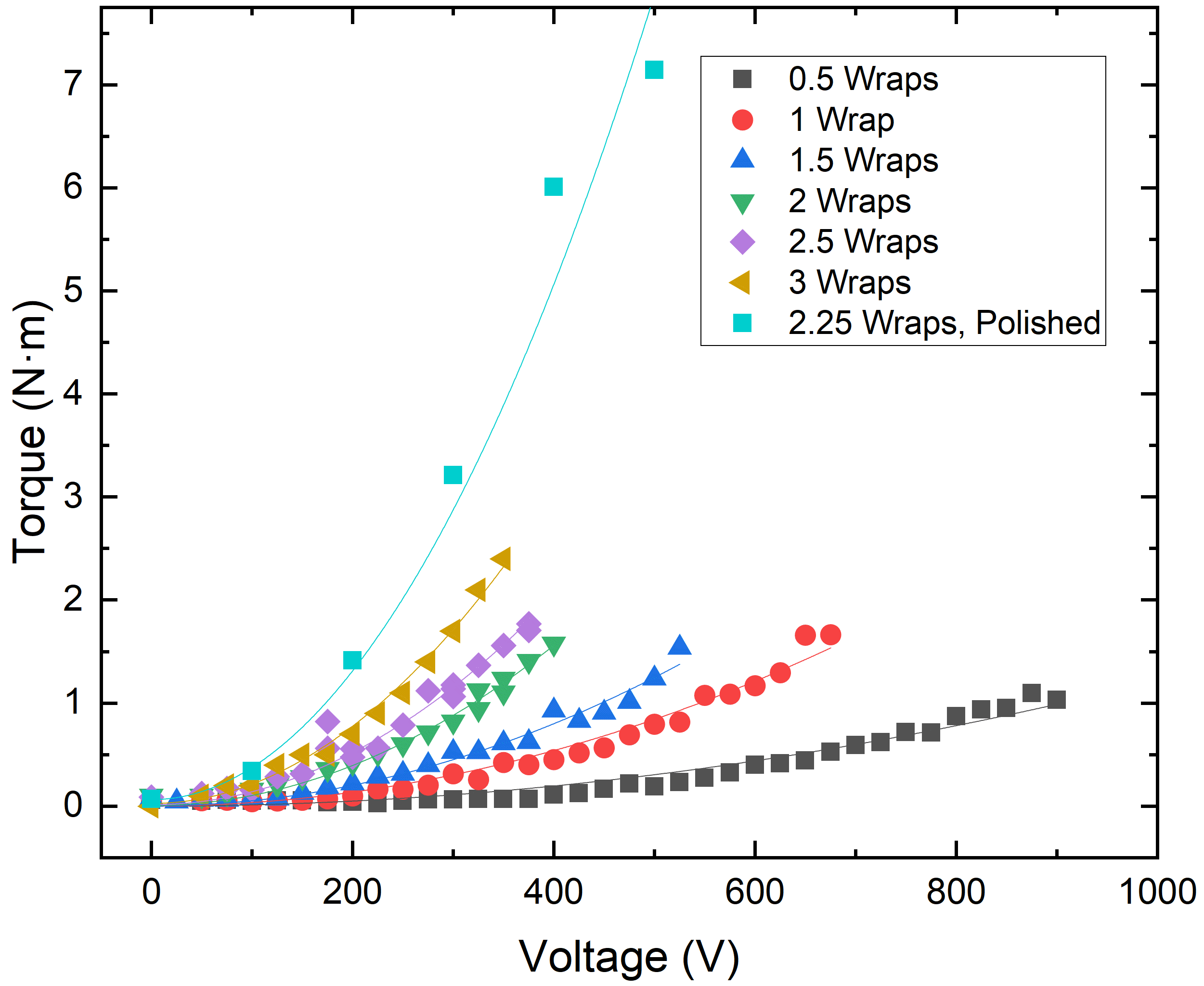}
 \vspace{-3mm}
 \caption{Data was fit to the derived JR-capstan equation assuming constant COF of $\mu$=0.20 with varying gap distance. Gap distance for 0.5, 1, 1.5, 2, 2.5, and 3~wraps were fitted to be 2.3~µm, 2.3~µm, 2.9~µm, 2.9~µm, 3.6~µm, and 4.1~µm, respectively, and 1.9~µm for 2.25~wraps polished band.}
 \label{fig:modelfit}
 \vspace{-0.75cm}
\end{figure}

To calibrate our model, several parameters are needed: the relative permittivity of the PBI, the coefficient of friction, and the gap between the dielectric and the braking band. The relative permittivity of 3.9 for PBI was determined using a vector network analyzer at 1~kHz.
To measure the coefficient of friction $\mu$ for the PBI material against a stainless steel shim, we used the standard capstan equation \eqref{capstan} since the holding tension is known and the output torque is measured. The experiments to determine $\mu$ were conducted using a $6\pi$ wrap angle. With the voltage off, we applied various holding tensions from 10~g to 130~g in approximately 10~g increments. An input torque was increased until the clutch slipped. The calculated coefficient of friction (COF) based on these experiments was 0.20.

For the 25.4~µm thick band a pretension $T_{hold}$ of 0.05~N was used. For the high torque version with a 76.2~µm thick band a larger pretension of 2~N was necessary to get the band to conform to the shaft. For data fitting to the 0~V state, a 0.3~N value was used to compensate. 

We measured the $T_{load}$ at which the clutch slipped for a range of voltages and wrap angles using both bands as shown in Fig.~\ref{fig:modelfit}. The air gap was used as a free parameter to fit the experimental data. These gap distances are reasonable for the material and system configuration. One consequence of increasing the wrap angle is that as the air gap distance increases, performance decreases. As wrap angle increases, it becomes practically harder to ensure that there is no twisting or lifting of the electrode edges. %Polishing the band surface was found to decrease the gap distance which will have a significant impact on the JR force by decreasing the air-gap between the wrapped electrode and dielectric. 

The proposed model in \eqref{gov-eqn} closely predicted the experimental data across multiple wrapping angles for the 25.4~\um band, with the correlation coefficients for 0.5, 1, 1.5, 2, 2.5 and 3~wraps respectively: 0.95, 0.97, 0.98, 0.98, 0.95, and 0.99. The 76.2~µm thick polished band correlation coefficient was 0.97 with the artificially decreased holding tension.

We see from Fig.~\ref{fig:modelfit} that increasing wrap angle significantly improves the holding torque generated by a clutch. Importantly, we also see that improving surface conformation improves clutch performance. The 2.25~wrap 76.2~\um band out performs the 3~wrap clutch at the same voltage. This is primarily due to the improved surface finish on the band. We also can note that the thicker a band is, the larger the ultimate holding torques that can be achieved. The 25.4~\um band fails at 2.5~\nm; however, the thicker 76.2~\um band can generate and sustain 7.1~\nm of torque.

%Gap distances for 0.5, 1, 1.5, 2.5, and 3 wraps was found to be 2.3 µm, 2.3 µm, 2.9 µm, 3.6 µm, and 4.1 µm, respectively. 

  %We should note that if µ = 0.15, range of gap distances narrows to 2-2.5µm and the fits are virtually unchanged.  

 % 

%With the clutch design there is a trade off between wrap count and voltage. 

\subsection{Effect of Pretension}

\begin{figure}[t]
 \centering
 \includegraphics[width=0.9\columnwidth]{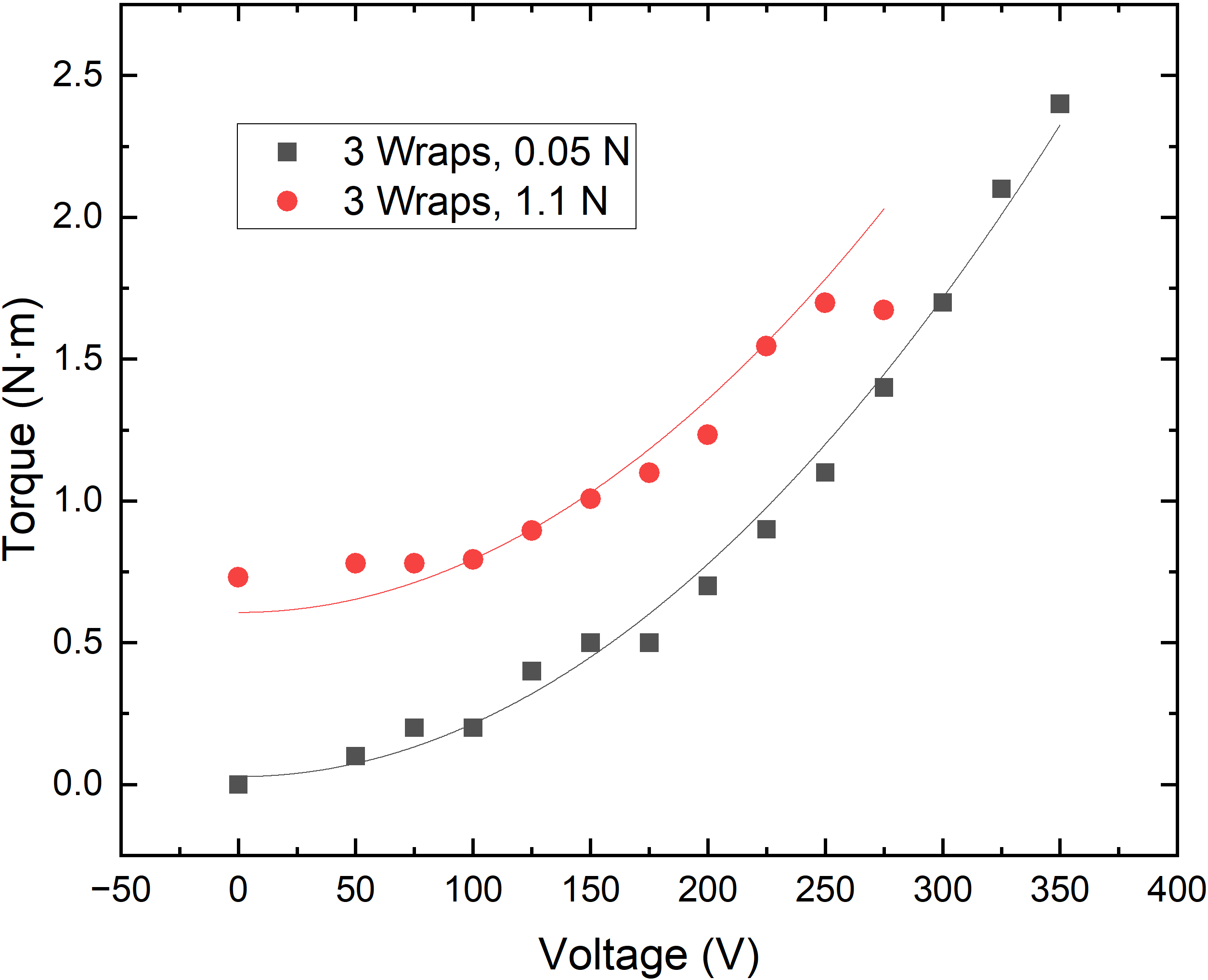}
 \vspace{-3mm}
 \caption{Effect of pre-tension on holding torque. Theoretical fit to model assuming equal gap of 4.1~µm and only varying input holding force $T_{hold}$. Pretension increases the holding torque but will also increase rolling friction (0~V operating point).}
 \label{fig:pretension}
 \vspace{-0.35cm}
\end{figure}

%\vspace{-0.30cm}
Fig.~\ref{fig:pretension} shows the effect of increasing the pretension from 0.05~N to 1.1~N on a 25.4~\um band with a wrap angle of 6$\pi$. When a lower pretension is used, the system has a negligible resistance of 0.1~N·m at 0~V. By contrast, the system has 0.75~N·m holding torque from a 1.1~N preload at 0~V. Increasing preload shifts the observed holding torque at a given voltage for an identical JRCC, but at a cost of requiring a higher 0~V holding torque. The lines in the chart represent the theoretical clutch model. We see close agreement between our model and the experimental data.
%Increasing the preload adds a dc offset, increasing holding torque at a given voltage for an identical JRCC but at a cost of a higher 0 V holding torque.

\subsection{Power Consumption}
\begin{comment}
\begin{figure}
 \centering
 \includegraphics[width=0.9\columnwidth]{Figures/Power.png}
 \caption{The power consumption of a JRCC with different wrap angles.}
 \label{fig:power}
 \vspace{-0.05cm}
\end{figure}
\end{comment}

\begin{figure}[t]
 \centering
 \includegraphics[width=0.9\columnwidth]{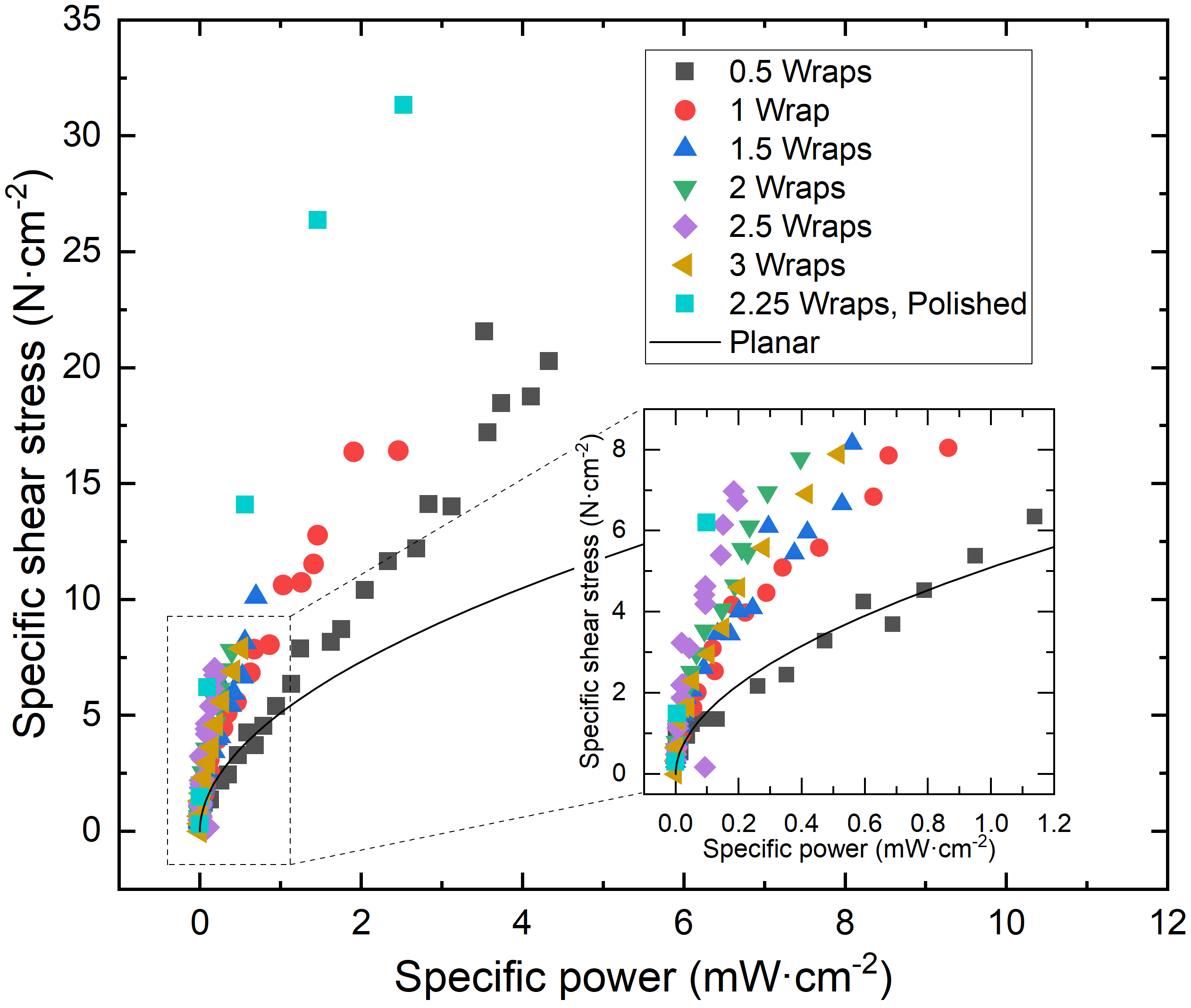}
 \vspace{-3mm}
 \caption{Power consumption vs. specific tension for various wraps. Higher wrap angles produce higher specific tension per specific power, making the device more efficient at larger wrap angles.}
 \label{fig:specificpower}
 \vspace{-0.75cm}
\end{figure}

\begin{comment}
    
Figure~\ref{fig:power} shows power consumption versus voltage for different wrap angles. Best seen from the $\pi$ wrapped case, there is a flat power draw until around 400 V. The wrapped electrode is made out of 316 stainless steel and is not formed to the shape of the central shaft. A certain amount of preload is required to have the electrode conform to the shaft. If the electrode is not conformed to the shaft the capstan equation no longer holds. This flat portion is from the clutch not engaging under limited preload. It takes a certain voltage until the band starts to adhere to the central capstan, enacting the capstan effect.
\end{comment}

Fig.~\ref{fig:specificpower} shows that increasing wrap angle increases specific tension per specific power, providing a metric on the efficiency of power conversion into holding torque. The increase in specific tension per specific power can be attributed to the capstan effect accounting for a portion of the total holding tension, which does not consume any electrical power \eqref{gov-eqn}. The planar case is modeled using the smallest gap measured for a 25.4~µm band (2.3~µm) and COF~µ~=~0.2. Compared to the modeled planar case, the power consumption for one or more wraps is significantly lower.

\section{Comparison of the JRCC to other clutches}

\begin{figure}
 \centering
 \includegraphics[width=0.9\columnwidth]{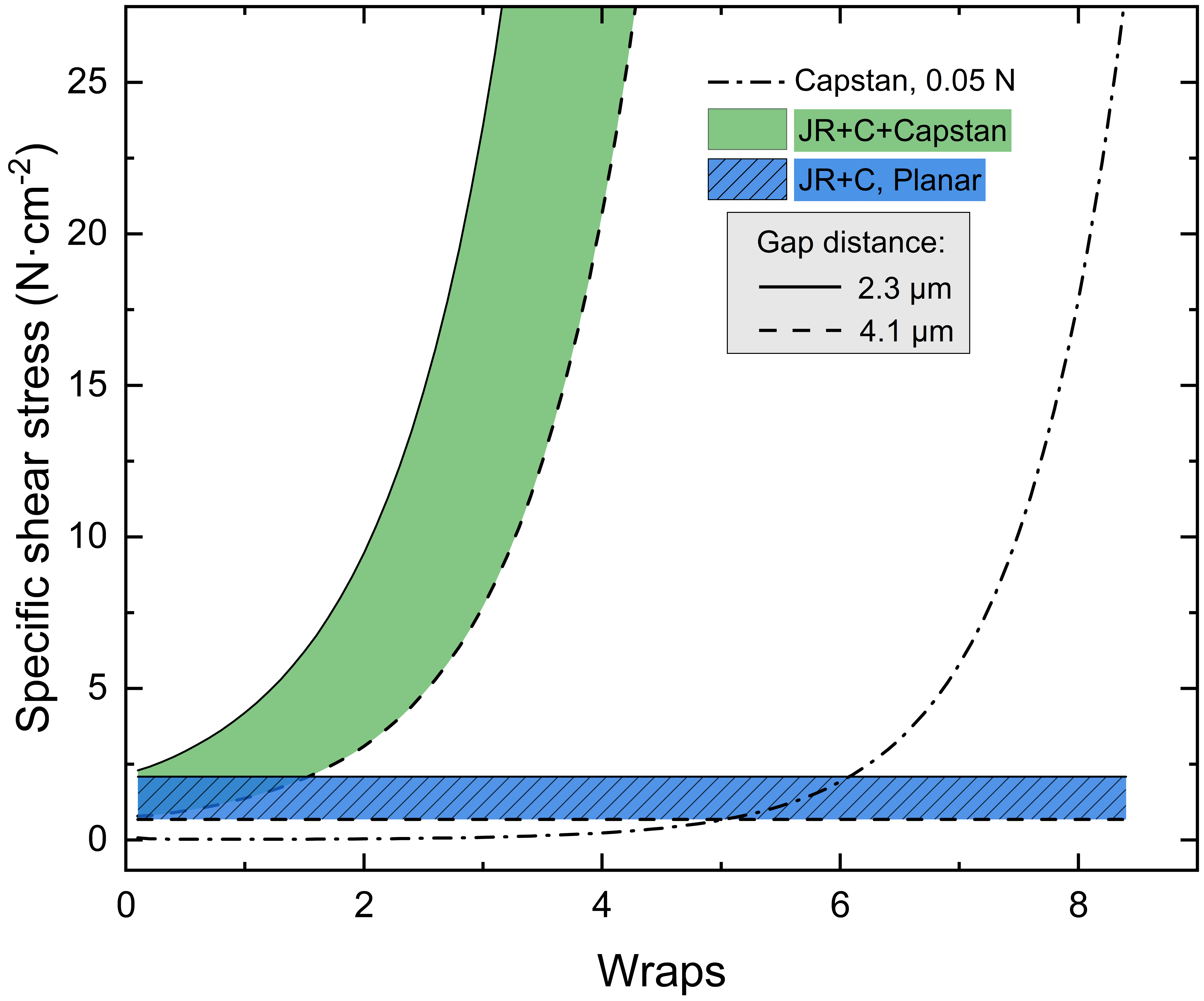}
 \vspace{-3mm}
 \caption[]{Comparison of various equivalent electrostatic clutch designs. The JR+C+Capstan (green) and JR+C (blue) region show the theoretical effect that gap distance has on the device. As seen here with the JRCC design, at higher wrap angles the impact of gap distance grows. Due to the capstan effect paired with JR and Coulombic electrostatic adhesion, our design is superior in specific shear stress and scales significantly faster with increased area.}
 \label{fig:model-comparison}
 \vspace{-0.75cm}
\end{figure}

%\vspace{-0.3cm}
We evaluated our design using the 25.4~\um band relative to theoretical benchmarks for other potential clutch designs. A capstan, planar equivalent, and a theoretical JRCC design along with experimental data is shown in Fig.~\ref{fig:model-comparison}. An air gap between 2.3 and 4.1~µm is shown to demonstrate the sensitivity to the gap variations observed in Fig.~\ref{fig:modelfit}. For a linear clutch, specific shear stress remains constant but increases for a JRCC due to the capstan effect, outperforming a planar design. Therefore the greater the wrap angle, the more efficient a JRCC will become. In Fig.~\ref{fig:model-comparison} we see how the different elements of our design interact. The JR and Coulombic effect can increase the specific shear stress of a clutch. However, the capstan effect dramatically increases the specific stress as the wrap angle increases. We can also see that while a planar design is sensitive to the gap between the band and shaft, the sensitivity of the JRCC design increases as wrap angle increases. This is shown by the width of the green region. We experimentally verified the increased sensitivity of the design by comparing the polished 76.2~\um band and the unpolished 25.4~\um band in Fig.~\ref{fig:comparison}. The improved performance from polishing is in line with the increased sensitivity to gap size seen in Fig.~\ref{fig:model-comparison}. 

We compared our JRCC design with both bands (Fig.~\ref{fig:comparison}) based on specific shear stress with clutches reported in the literature~\cite{Patrick-2020-underactuated,Diller-2016-clutch,Diller-2018-DesignParam,Hinchet-2020-High-force,Hinchet-2018-WearableBrake,Karagozler-2007-Oragami,Ramachandran-2019-Textile,Wang-2019-Variable-Stiffness}. We see that using the capstan effect allows for increased specific shear stress for the same material. This is shown in the mapping of the planar case in the blue dashed line to the region of the 25.4~\um band highlighted in red. Our design with the 76.2~\um band generated a specific shear stress of 31.3~N/cm$^{2}$, the highest value currently recorded in the literature. Compared to the previous state of the art~\cite{Hinchet-2020-High-force}, our JRCC with the 76.2~\um band required a higher voltage of 500~V vs. 300~V and consumed more power at 2.5~mW/cm$^{2}$ compared to 1.2~mW/cm$^{2}$. In theory, if we were to use a more effective dielectric material such as that reported in \cite{Hinchet-2020-High-force}, and combine it with our design, even greater specific shear stresses and more efficient clutches could be achieved.

\begin{figure}[t]
 \centering
 \includegraphics[width=0.9\columnwidth]{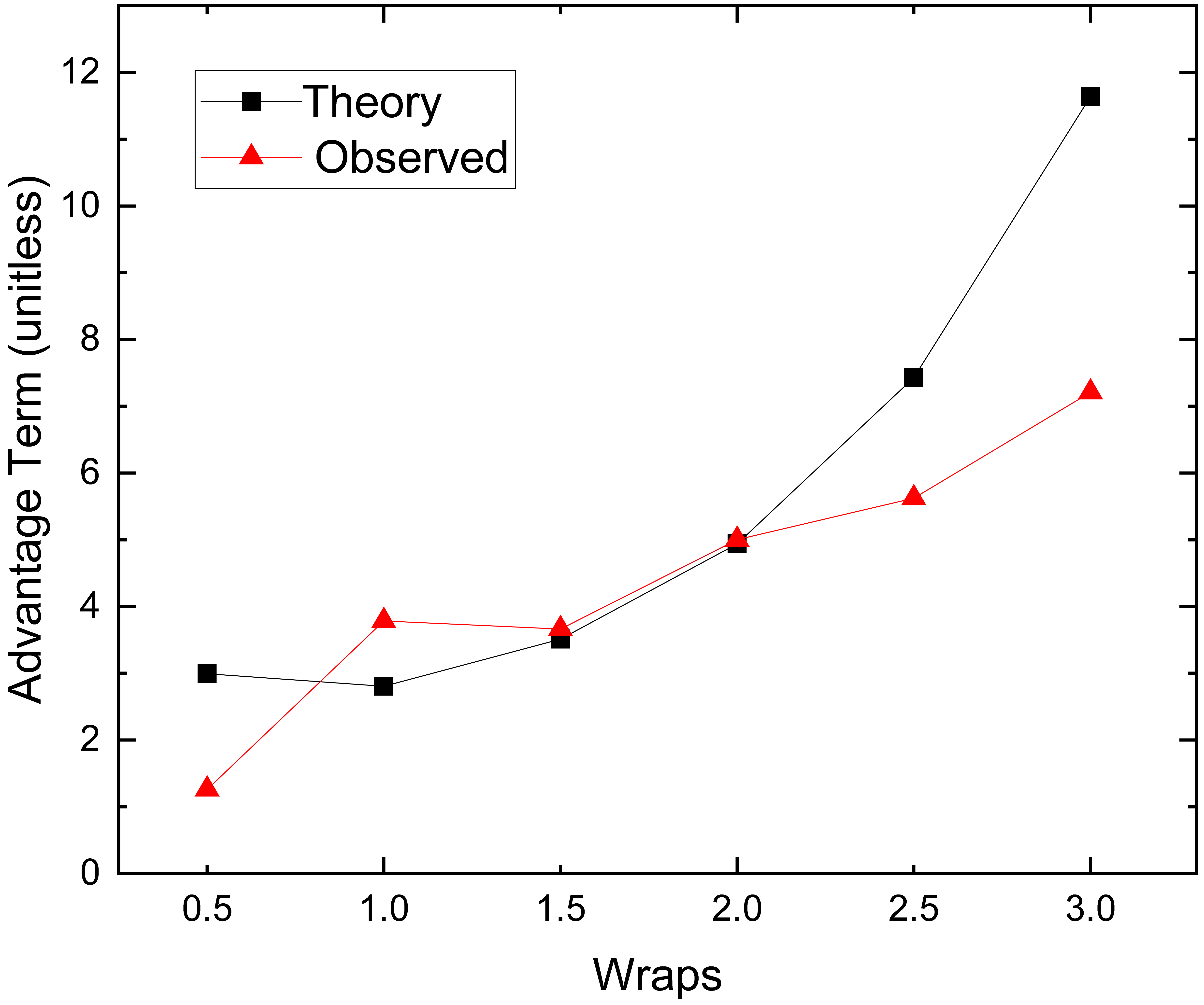}
 \vspace{-3mm}
 \caption{Theoretical advantage compared to experiment on 25.4~\um band. %This model is evaluated 350 V using the average air gap of 3.2 µm across all the experimental wrap angles. The model used a band width of 10 mm, a thickness of 25.4 $\mu$m and µ of 0.20
 }
 \label{fig:adv}
 \vspace{-0.6cm}
\end{figure}

In the JRCC design, force per unit area scales with $(e^{\mu\theta}-1)/(\mu\theta)$, the capstan advantage term. For the larger wrap angles, the higher portions of their exponential curves \eqref{gov-eqn} are realized. Fig.~\ref{fig:adv} compares the calculated advantage term \eqref{adv-eqn} of our clutch design using the 25.4~\um band at 350~V, $\mu$ of 0.2, and calculated air gaps used for data fitting as shown in Fig.~\ref{fig:modelfit}. The advantage term aligns nicely with theory but falls off at higher wrap angles. As discussed in section \ref{sec:ExpCharacterization}, increasing wrap angle is functionally harder to construct and has a practical loss in performance that can be fixed with better materials and fabrication procedures. Fig.~\ref{fig:adv} demonstrates that no matter what dielectric is used, a capstan-based design will always have an advantage over a typical planar design.

% We also compared our designs to the theoretical output torques a planar clutch reported from the literature could produce on our 1 inch diameter shaft. We see in Figure ~\ref{fig:torque-comparison} that out capstan design out performs all of the other structures. \TA{Talk about the Data and what we see here} ~\ref{fig:modelfit}. Therefor we can see how the multi-wrap capstan design is a much more efficient clutch at converting voltage to torque. Therefor we can take a lower performing dielectric material, and use the capstan effect to make it more efficient at producing torque than a standard planar clutch and produce using the most performant materials to date. 

\vspace{-.1cm}
\section{Conclusions and Future Work}
\vspace{-.2cm}
We constructed an electrostatic clutch demonstrating 31.3~N/cm\textsuperscript{2}, the highest shear stress in the literature to date. This was possible by combining the capstan effect with JR and Coulombic electrostatic adhesion. Our device delivered a holding torque of 7.1~N·m on a 25.4~mm diameter output shaft using only 500~V and consuming 2.5~mW/cm$^{2}$. Additionally, we provide a design framework and present data demonstrating the accuracy of our model. Using this model, we predicted that for equivalent geometries and materials, a JRCC will outperform any planar construct \eqref{adv-eqn} as shown in Fig. \ref{fig:adv}. Our design can be used to generate clutches with higher holding forces, enabling new applications. 

The current implementation of the JRCC is limited by both the mechanical properties of the stainless steel band and the growth in gap size as wrap angle increases. After the initial exponential pattern was verified, a thicker band was utilized to withstand larger shear stresses. This required surface polishing and a larger holding torque to compensate for the increase in rigidity, which moved further from approximations made in the derivation of \eqref{gov-eqn}. Future work will explore the best models and materials for a capstan band that can conform to the dielectric substrate while withstanding sufficient shear stresses. Future work will also explore designs that allow motion in multiple directions.

%This design was also evaluated using polyimide (PI) film attached to the central capstan. Interestingly besides needing much higher voltages (>1000V), the response was linear. Looking at the coefficient of friction data it looks to follow power law friction rather than Coulomb friction. This makes the response much more linear, similar to what is shown here [Heap 2023, Capstan Winch] when the capstan band obeys power law friction.
\vspace{-.25cm}
\section*{ACKNOWLEDGMENT}
\vspace{-.2cm}
The authors thank Dr. Alex Langrock for assisting in the identification and acquisition of PBI for this work. This work was funded in part by a gift from the UW + Amazon Science Hub and DEVCOM ARL CRADA 19-005-J003. A provisional patent has been filed on the design. The views and conclusions contained in this document are those of the authors and should not be interpreted as representing the official policies, either expressed or implied, of the Army Research Laboratory or the U.S. Government.

%\TODO{Is this okay as acknowledgement for ARL?}

%%%%%%%%%%%%%%%%%%%%%%%%%%%%%%%%%%%%%%%%%%%%%%%%%%%%%%%%%%%%%%%%%%%%%%%%%%%%%%%%
\bibliographystyle{IEEEtran}
\bibliography{refrences}

\end{document}